# A Machine Learning Framework for Real-Time Personalized Ergonomic Pose Analysis

**Manex Atxa**, Bruno Simoes, and Julen Balzategui

*Vicomtech Foundation, Basque Research and Technology Alliance (BRTA),*

*Mikeletegi 57, 20009 Donostia-San Sebastián (Spain)*

This paper introduces a new methodology for real-time prediction of ergonomic and non-ergonomic human poses using volumetric video data in three dimensions. Although the methodology was designed for ergonomic assessments, it can be adapted to other applications requiring real-time analysis of human posture. One aspect that makes this system stand out is its ability to analyze 3D point clouds during the assessment, enabling computation from multiple angles. This overcomes a critical limitation of cameras which provide often a fixed viewpoint, thereby restricting the data available for a thorough postural evaluation, especially when occlusions occur. The system continuously and automatically performs pose inference using the chosen perspective on the real-time streaming data; however, only the poses manually selected and labeled by the user are used to train the personalized deep learning classifier. The methodology has been refined through a case study in which RGB-D cameras captured subjects performing load-lifting tasks, enabling real-time skeletal labeling. The model was trained on this data and, following the training phase, performs inference on new streaming data in real time. This research offers a scalable and pragmatic approach for real-time ergonomic evaluation by combining state-of-the-art 3D data technologies and traditional 2D pose estimation algorithms. It addresses the increasing need for safety and health monitoring in workplace environments, marking a notable contribution to the domain.

## 1. INTRODUCTION.

The application of artificial intelligence (AI) technologies in ergonomic assessments has substantially improved the ability to assess and manage workplace safety concerns. AI technology and sophisticated ergonomic assessment techniques have yet to become established in most industry sectors and organizational structures. Current techniques rely mainly on observational scoring systems that utilize manual scanning, which has problems such as time-consuming nature, inter-rater reliability issues, and limited scalability for large workforce populations [1]. Solutions powered by AI, particularly those based on computer vision systems and wearable sensor networks, offer automation through sophisticated data collection and processing and are superior concerning accuracy, efficiency, and real-time monitoring for ergonomic risk assessment. In addition, a recent exploratory study [2] on AI-enabled ergonomics analyzed primary factors influencing adoption and implementation barriers through surveys. Findings suggested that it is possible to mitigate the operational risks and costs while improving efficiency through ergonomic innovations.

The proposed system combines 3D volumetric video data with traditional 2D pose detection algorithms allowing for real-time ergonomic evaluations. Real-time detection for ergonomic applications demonstrates reduced computational intensity compared to other real-time systems, as transitions between ergonomic and non-ergonomic poses occur infrequently within single-second intervals due to inherent human movement limitations. Along with this approach comes the benefit of enabling the users with 2D pose detection interfaced within a responsive 3D point cloud framework. The users are able to rotate, zoom in and out, and also shift the viewpoint of the point cloud data allowing for intuitive analysis of human posture from multiple angles. In traditional approaches a fixed perspective is used to obtain image captures and while this makes pose interpretation easier, the data available for analysis is limited, in case of occlusions. Proposed systems works with data captured by an array of RGB-D cameras which allows the direct generation of complete 3D point clouds resulting in numerous virtual perspectives capturing rich detail for enhanced ergonomic assessment. For detecting ergonomics in 3D point cloud data, a two-dimensional projection is performed through depth buffer acquisition, enabling the application of established 2D pose estimation methodologies, which demonstrate greater effectiveness, reduced computational requirements, and enhanced accuracy in real-world implementation scenarios.

LiDAR technology has transformed construction site safety by enabling real-time monitoring and quality control in 3D data capture [3]. The incorporation of 3D data with AI-powered ergonomic evaluations proves to be exceptionally beneficial in multiple industries. In office settings, 3D point clouds resulting from RGB and depth imagery are processed alongside 2D algorithms for assessing worker posture [4]. These advancements highlight the need for integration of 3D data within ergonomics, especially in environments with many people, changing tasks, and various requirements. Though 2D detection methods have largely superseded advanced techniques, retrieving the spatial configuration of a target's body parts directly from a point cloud is still a challenge due to the need for clear access to the point cloud data as well as the efficiency of streamlined step-by-step information processing. Hence, the technique proposed in this paper concentrates on optimizing the implementation of pose detection in 2D within a given framework, attaining an equilibrium between strong functionality, extensibility, and the thorough detail of spatial analysis allowed by the 3D point cloud interface.

## 2. LITERATURE.

Several studies have explored the applicability of artificial intelligence within the field of ergonomics, with some reviews covering its application for safety within the workplace, human behavior analytics, and injury forecasting [5]. AI methodologies own capabilities such as handling big data datasets [6], observing human postures and movements along with their interactions which are associated with the musculoskeletal risk factors. The ergonomic assessment approaches powered with AI can be systematically organized into two main groups: posture and movement recognition models based on computer vision [7], and those based on sensors which use biomechanical data obtained from devices such as surface Electromyography (sEMG) sensors [8].

Compared to computer vision models, sensor-based models require an elaborate set of hardware equipment, direct interaction with the user, and have implementation problems such as intrusiveness. For this reason, our work employs computer vision techniques using RGB-D cameras for remote data collection. Such techniques offer a non-intrusive, adaptable and user-friendly approach which avoids the need for wearable devices enhancing user experience and ease of implementation.

Models within the realm of computer vision are capable of processing both two-dimensional [9] and three-dimensional [10] data. RGB-D cameras capture both 2D imagery and depth information concurrently. This permits more comprehensive scene understanding and improves object detection, gesture recognition, and spatial mapping. These models can therefore be organized according to the dimensions of data processing.

Two-dimensional input-based models operate with image data obtained from planar views using standard RGB cameras. They extract colors, textures, edges, and patterns to interpret images visually. Such systems are integrated with neural networks allowing for automatic detection and classification of body postures which can expedite ergonomic risk evaluations in industrial settings. The described application uses a 2D pose detector based on an open-source framework called MMPose [11], which has demonstrated efficient speed and accuracy in the COCO keypoint detection benchmark [12].

These methodologies have changed significantly within different paradigms. Approaches are classified based on several criteria. Research may center around single-person pose estimation [13] while other works focus on algorithms for multi-person pose detection [14]. Early techniques mostly relied on coordinate regression predicting joints' coordinates as outputs for the given images input [15]. Further advancements developed the representation of heatmaps [16]. More recent works started to explore different novel architectures such as those with Transformer models [17] that apply Transformer network architecture [18] and GCN (Graph Convolutional Network) based models [19]. All these models achieved encouraging results for human pose estimation. Nevertheless, today's algorithms for 2D detection do not handle point clouds or 3D information as input.

Models using three-dimensional inputs operate on depth-enhanced data in the form of point clouds, which convey the spatial configuration of objects and environments. This enables the recognition of their shape, depth estimation, and spatial reasoning [20]. Technologies such as LiDAR and stereo camera systems offer 3D point clouds which, when coupled with AI methodologies, allow for evaluating ergonomic risks by combining body posture within the context of movement spatially. Thus, volumetric video alongside point cloud processing has become pivotal in ergonomics, construction, and robotics. For instance, in recent studies, LiDAR-based systems are being used to monitor workers' movements and interactions with machinery in real-time, helping to identify hazardous postures and ensure compliance with safety standards [3]. Similarly, point cloud data can be combined with 2D pose detection algorithms to track human movements in 3D space, offering a comprehensive view of the individual's posture and the surrounding environment [4]. Nonetheless, inferring pose detection directly from point clouds is not as robust as state-of-the-art 2D detection, as point clouds require clear and unobstructed data or sequences to function properly [20, 21]. In contrast, 2D pose estimation algorithms like OpenPose [16] are well-optimized and trained on large, diverse datasets such as COCO [22].

Consequently, the presented application addresses the previously mentioned drawbacks. Firstly, it allows 3D data as input, enabling the handling of an interactive point cloud with infinite view perspectives. Additionally, it projects the 3D data onto a 2D space, enabling the application of robust, state-of-the-art 2D pose estimation methods.

## 3. SYSTEM ARCHITECTURE.

This section describes the two-part system which begins with inputting point cloud data and culminates with obtaining classified skeletal structures. The first component involves training a deep learning model for pose classification, while the second component details the data processing steps for model inference.

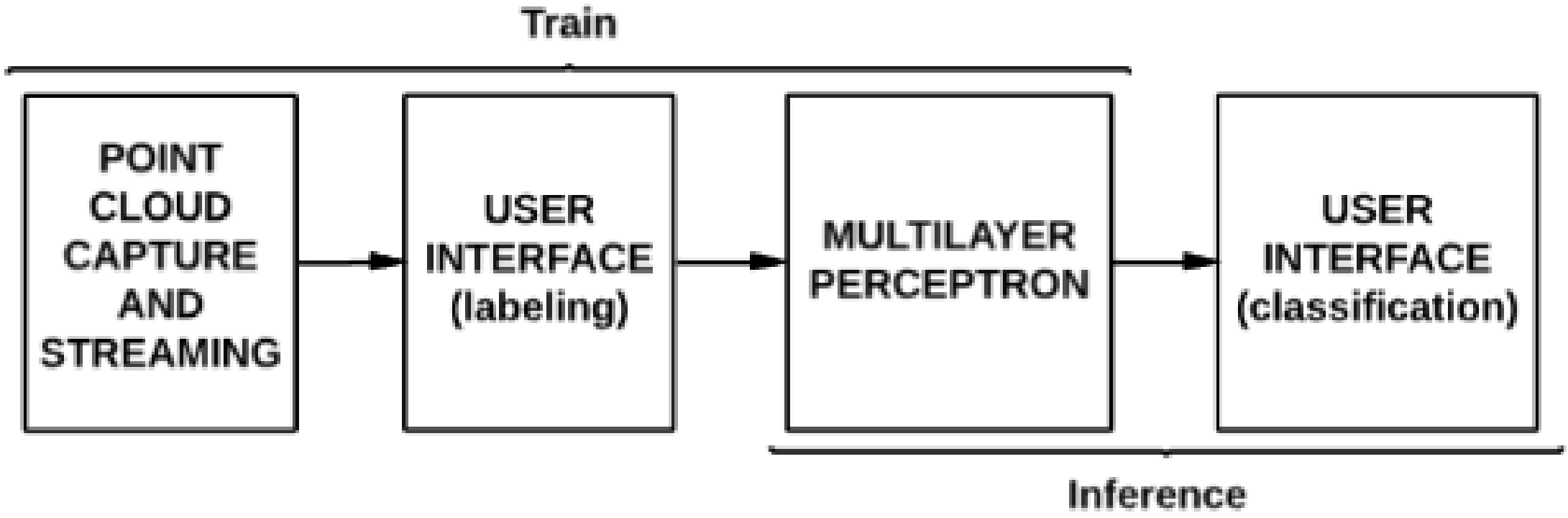

***Figure 1:*** *Two-Part System Architecture.*

### 3.1. TRAINING STAGE.

**Figure 2:** *Training Pipeline.*

The RGB-D cameras are configured as a multi-camera system. Point cloud data are reconstructed utilizing the system outlined in [23] and sent over WebSockets for real-time processing. A WebSocket server is part of a microservice-based architecture, configured to allow the streaming of point cloud data towards an AI module which needs to process the data and perform pose detection. The transmission of point cloud data using a WebSocket-based communication protocol is essential for applications such as real-time ergonomic pose estimation.

When the streaming of point cloud data is initiated, point clouds are rendered in an interactive interface where users can perform rotations, translations, and zoom in and out an unlimited number of times. Even while pose detection algorithms in real time are applied to the image at hand, the point clouds are projected onto two-dimensional space by the GPU. A 2D pose detection algorithm analyzes the two-dimensional projection of the point cloud, and detects the keypoints on the person. If detection fails due to occlusions, the user can feed the system with a better viewpoint of the 3D scene.

As skeletons are being detected, the user has the ability to annotate the skeleton interactively on the screen using a set of labels, which can be tailored to specific requirements for a given project. This feature allows the formation of labeled datasets of the skeletal shapes, which are composed of the coordinates of the keypoints that have been computed. These annotated instances are retained until the user has finished the annotation session. After a dataset is finalized, that dataset can be employed to teach a Multi Layer Perceptron (MLP) classifier to classify new data based on the labeled sparse skeletal information (Figure 3).

The x and y coordinates of each keypoint serve as inputs for two neurons at each keypoint for the MLP classifier's input layer. Therefore, the trained model will have an input layer containing twice the number of keypoint neurons. Before training, the user specifies the parameters for the MLP classifier, including the number and size of hidden layers, the maximum number of iterations, the training and testing data sizes, the solver algorithm, and the activation function to be used. Once these parameters are set, the training process begins, and the output is a trained MLP classifier capable of categorizing incoming skeletons into the predefined labels.

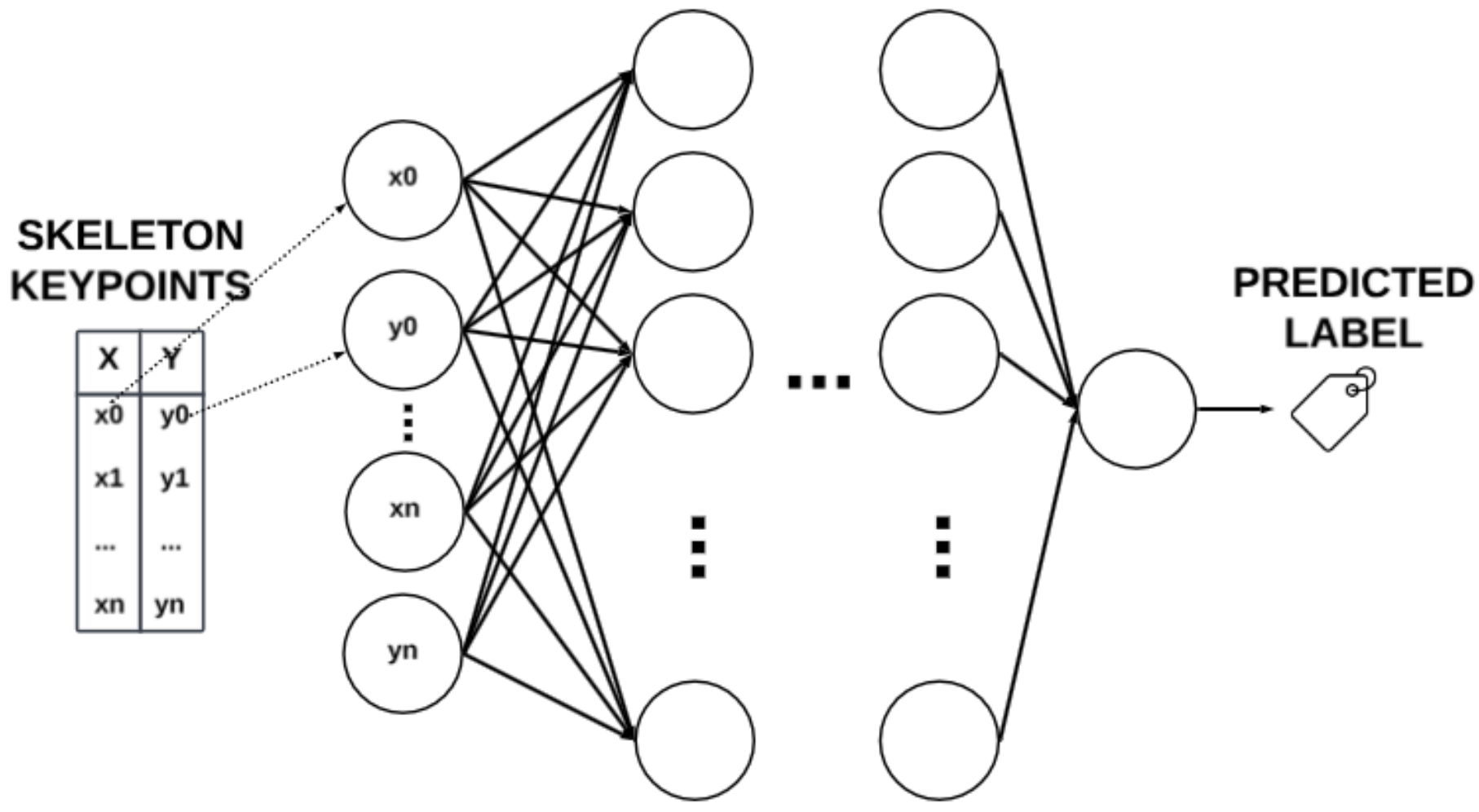

***Figure 3:*** *Multi-Layer Perceptron Architecture and Functionality.*

### 3.2. INFERENCE PHASE.

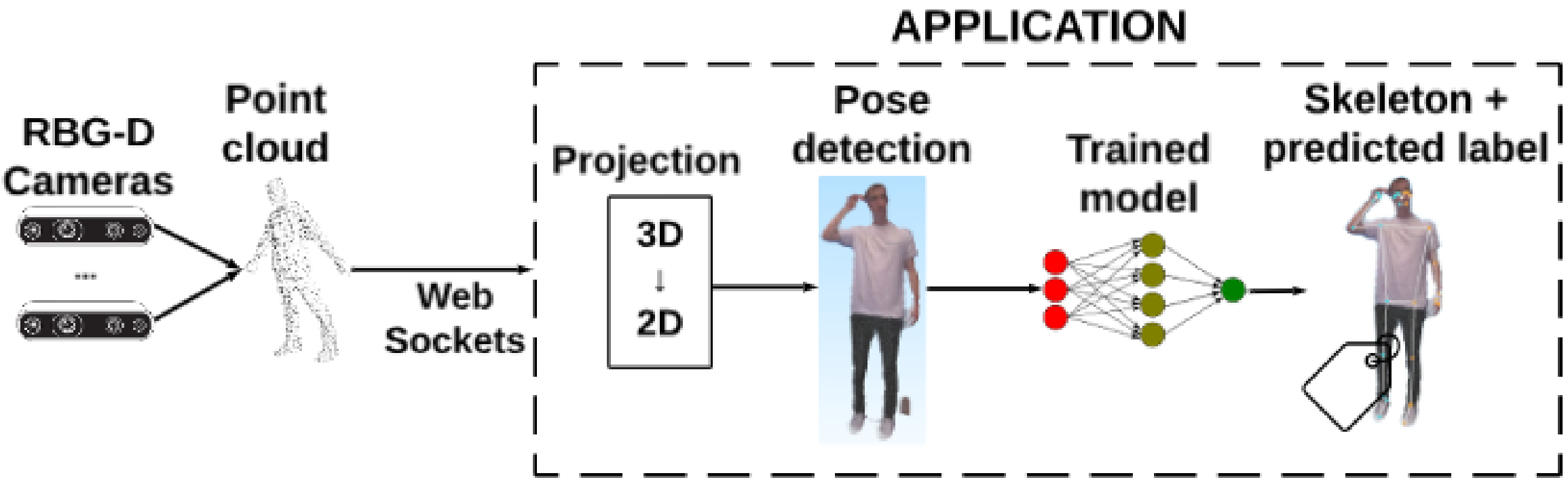

***Figure 4:*** *Inference Pipeline.*

After completing the training with Multi-Layer Perceptron, the trained model is loaded, and the system is ready to accept additional point cloud data through WebSocket interfaces. The process begins by accepting new point cloud data which is rendered in an interactive display and transformed into 2D. As in the training phase, skeletons are detected in real time, and the classifier processes them as keypoint coordinates to predict their labels with minimal latency.

The system output includes real-time data streams, detected 2D poses, and predicted classification labels, all displayed within the interface for user feedback and system response visualization.

## 4. USE CASE.

A specific use case was created to assess the functioning and effectiveness of the system. The aim was to categorize the poses of workers who were load lifting as either ergonomic or non-ergonomic. In achieving this aim, the previously described architecture was adopted for the given scenario devised for the experiment. First, a custom dataset was developed that included exemplar datasets of both pose categories based on certain predefined ergonomic benchmarks outlined in the medical literature [24]. This specific dataset was then used to build a classifier that is capable of detecting and classifying poses for analysis of future data.

The experimental setup included two RGB-D cameras, which have the capabilities of capturing image color as well as depth information simultaneously. These cameras were placed strategically to obtain detailed colored point clouds that provided adequate spatial coverage. The cameras underwent specific calibration procedures to obtain measurements for depth, color, and other parameters so that high-quality three-dimensional data could be acquired.

As shown in Figure 5, two Azure Kinect DK cameras were mounted to form a dual camera RGBD system and were configured to perform RGBD video recording of subjects engaging in ergonomically and non-ergonomically correct cardboard box lifting exercises. Each Kinect DK includes a depth sensor of 1 MP, an RGB camera of 12 MP, and an array microphone of 7 channels which allows for capturing depth information, images of high resolution, and sound with the required level of precision. A wide variety of load-lifting techniques were captured which included all possible lifting postures. The diversity of this dataset made it possible to conduct a thorough analysis to differentiate between ergonomic and non-ergonomic poses which can be used for classification and later for an ergonomic analysis.

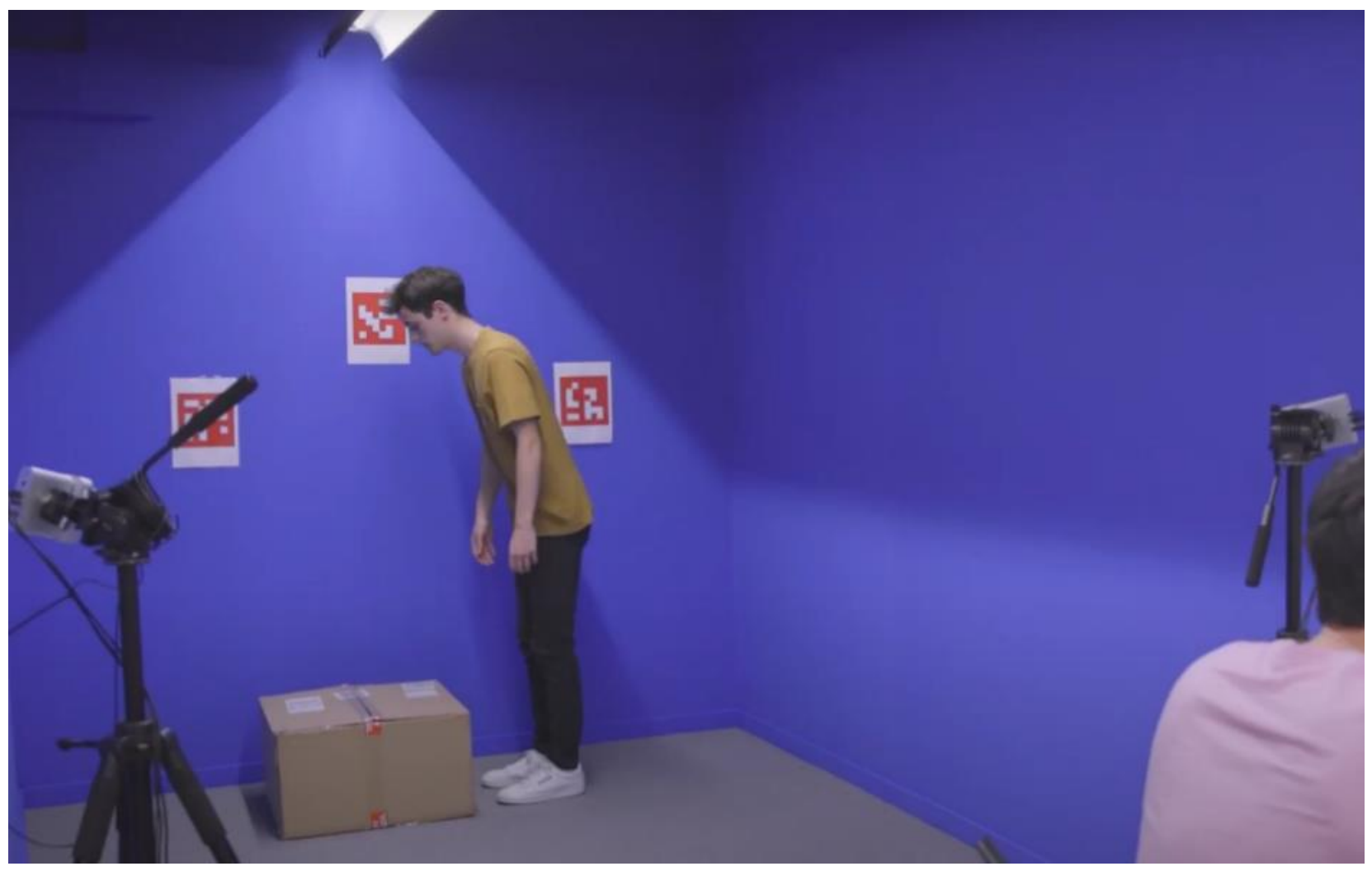

***Figure 5:*** *Pose Detection Set-Up.*

The 2D pose detection toolbox of the pose detection system MMPose [11] was used as the basis for the system's frame. Users have the ability to assign different labels, for instance, *ergonomic* or *non-ergonomic*, through labeling buttons on the interface as illustrated in Figure 6. Skeleton detection is carried out in real-time as users segregate skeletons, so timestamped selections can be made alongside the relevant skeleton keypoints to form the training dataset. Once a certain amount of labeled instances are gathered, users are able to stop this process and move straight to the training of deep learning models.

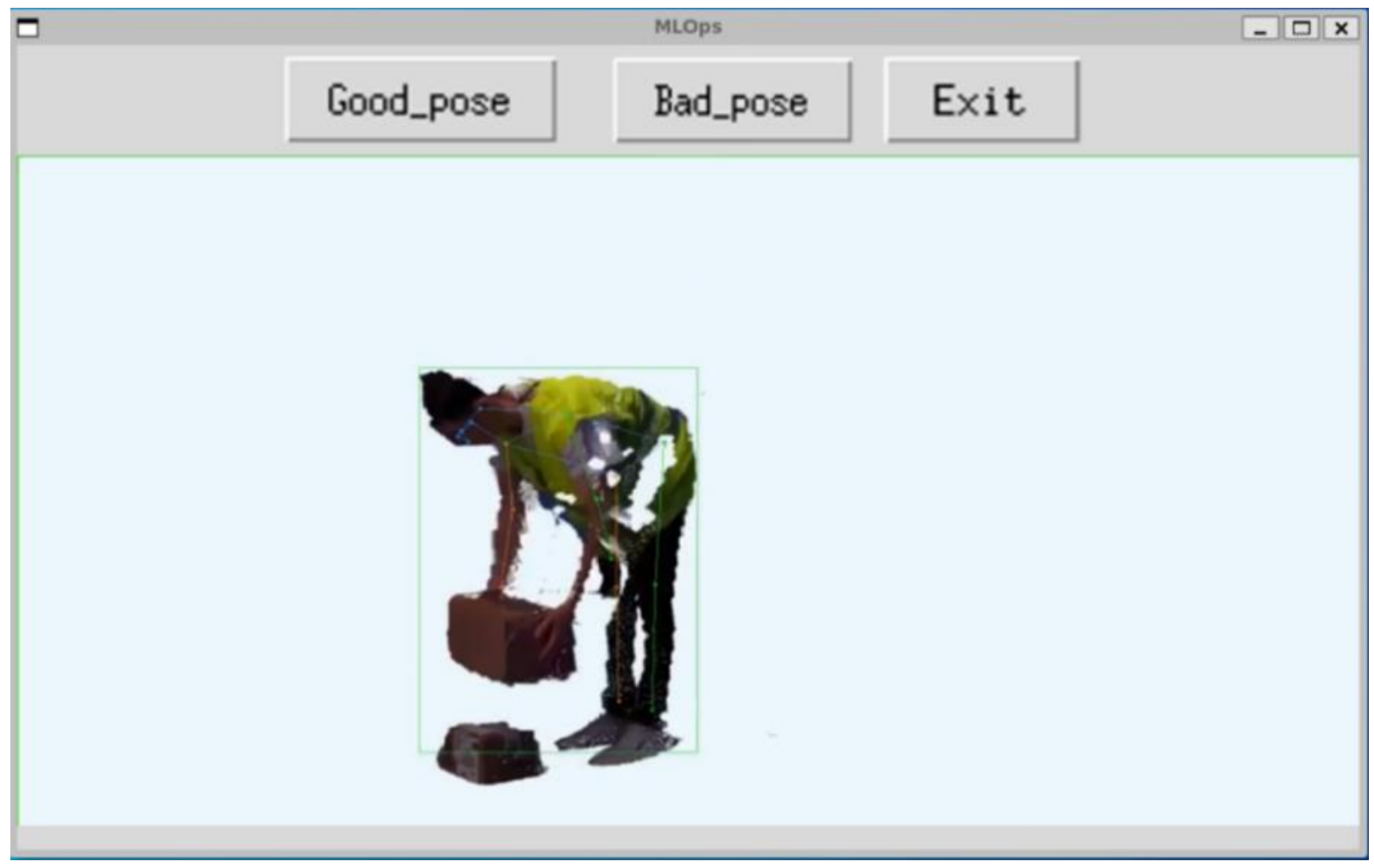

***Figure 6:*** *User Labeling Interface.*

Using Scikit-learn [25], a Multi-Layer Perceptron (MLP) classifier was developed. This library simplifies the development of efficient classification models, with only a few parameters requiring adjustment to achieve optimal results. Parameters are already preset through an application interface, which is convenient for many use cases.

In this case, the ratio of training to testing datasets was set at 80% to 20%. The architecture of the MLP classifier contained three hidden layers with 20, 10, and 7 neurons respectively. The training procedure was set to a maximum of 500 iterations using the ReLU activation function and the Adam optimizer.

Parallel computing tasks were executed on an NVIDIA GeForce RTX 3060 GPU and an Intel(R) Core(TM) i5-10400F CPU running at 2.90 GHz.

## 5. RESULTS.

As a result, we achieved real-time processing of new point clouds, displaying the input point cloud projected in two dimensions, the detected skeleton in both 2D and 3D—reconstructed using the capabilities of MMPose—and the predicted label, as shown in Figure 7.

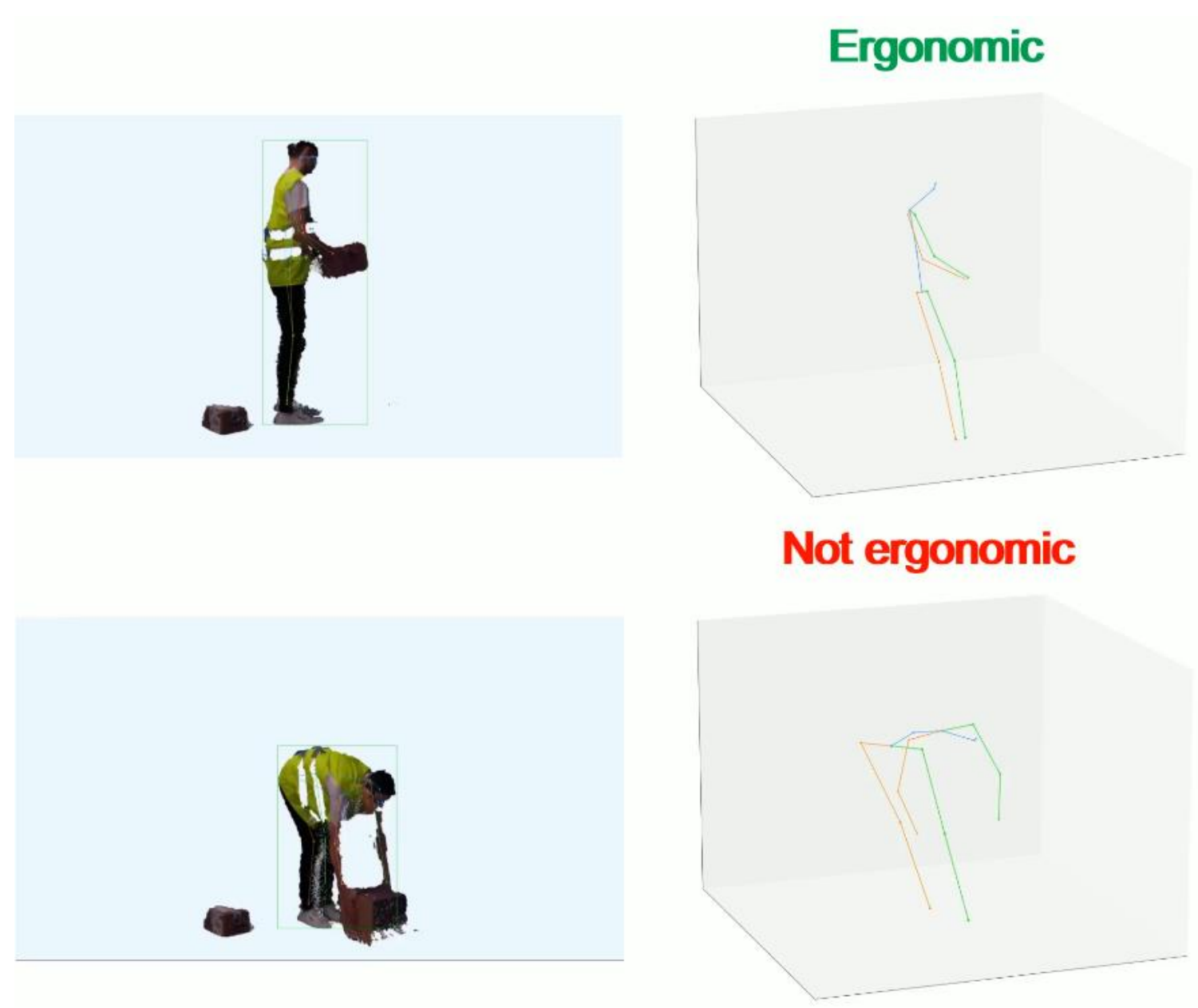

***Figure 7:*** *Ergonomic and non-ergonomic pose inferences.*

Within this study, time to completion is framed as a time-critical task, owing to the need to perform human pose detection and cloud data processing in a timely fashion within a pointed time span. In this case, the speed of processing is directly linked to the point number of the point cloud and the frame rate of the data stream. Table 1 displays the corresponding execution times for the pose as detected under different predefined conditions.

| Point Cloud Size | CPU (s) | GPU (s) |
|---|---|---|
| 1,000 | 0.36 | 0.12 |
| 15,000 | 0.40 | 0.13 |
| 50,000 | 0.43 | 0.17 |
| 120,000 | 0.47 | 0.22 |
| 300,000 | 0.62 | 0.37 |
| 475,000 | 0.81 | 0.52 |
| 750,000 | 0.98 | 0.75 |

***Table 1:*** *Execution time per detected pose for different point cloud sizes.*

These findings verify that the time taken to process data correlates with the increase in size of the point cloud data. Nevertheless, even when dealing with extensive point clouds, the system stays within tolerable limits for performance in real-time execution.

This system equilibrium strengthens interaction fluidity and accurate pose detection functionality.

The frame rates achieved for different point cloud sizes indicate the network processing capacity. As the point cloud size increases, the frame rate decreases, reflecting the greater processing demands on the system. Despite this, the system remains capable of maintaining a reasonable frame rate for real-time performance, especially for smaller point cloud sizes. It is important to note that the maximum frame rate allowed by the visualization system is 30 FPS, which is the upper limit for smooth interaction. Consequently, the application maintains 30 FPS when processing point clouds of 15,000 and 50,000 points, but the frame rate drops to 15 FPS for point clouds of 331,000 points.

## 6. CONCLUSION.

This paper describes an advanced machine learning system which combines volumetric video data, ergonomic pose analysis, and two-dimensional pose detection in real time. The system employs point clouds to capture and encode 3D spatial representations of objects within an environment. When integrated with robust computer vision methods such as 2D pose estimation, it enables accurate classification of ergonomic and non-ergonomic poses in dynamic work settings. Usability is enhanced via an interactive interface that permits users to manipulate 3D point clouds in real time while performing inference and labeling 2D skeletal data. Skeleton detection runs automatically and continuously, but only user-selected poses are saved to train a personalized classifier.

The verification of the method was done through a use case implementation with stereo camera systems, capturing workers in real-life environmental conditions. The experimental results confirm the capability of the method to detect ergonomic risk factors, proving its usefulness as a tool for safety evaluation in the workplace. Importantly, this analysis can support occupational health and safety by enabling early identification and mitigation of ergonomic risks, helping to reduce workplace injuries and improve worker well-being. Further refinement of the system will be pursued in other real-world scenarios, which include the application of advanced machine learning techniques and testing in more complex environments.

## 7. REFERENCES.